\documentclass[journal]{IEEEtran}
\IEEEoverridecommandlockouts
\ifCLASSINFOpdf
\else
   \usepackage[dvips]{graphicx}
\fi
\usepackage{url}

\usepackage{cite}
\usepackage{amsmath,amssymb,amsfonts}
\usepackage[ruled,vlined]{algorithm2e}
\usepackage{graphicx}
\usepackage{textcomp}
\usepackage{xcolor}
\usepackage{float}
\usepackage{multicol}
\usepackage{mathtools}
\usepackage[export]{adjustbox}
\usepackage{url}
\usepackage{multicol}
\usepackage{mathtools}
\usepackage{mathalfa}
\usepackage{colortbl}
\usepackage{breqn}
\usepackage{csquotes}
\usepackage{subcaption,graphicx}

\begin{document}

\title{Resource Efficient Perception for Vision Systems}


\author{A V Subramanyam,~\IEEEmembership{Member,~IEEE,}
        Niyati Singal,
        and Vinay K Verma
\thanks{A V Subramanyam, Niyati Singal, and Vinay K Verma are with the Indraprastha Institute of Information Technology Delhi (IIIT-Delhi), New Delhi, India (e-mail: subramanyam@iiitd.ac.in; niyati17032@iiitd.ac.in; vinayv@iiitd.ac.in).}}







\maketitle
\begin{abstract}
 Despite the rapid advancement in the field of image recognition, the processing of high-resolution imagery remains a computational challenge. However, this processing is pivotal for extracting detailed object insights in areas ranging from autonomous vehicle navigation to medical imaging analyses. Our study introduces a framework aimed at mitigating these challenges by leveraging memory efficient patch based processing for high resolution images. It incorporates a global context representation alongside local patch information, enabling a comprehensive understanding of the image content. In contrast to traditional training methods which are limited by memory constraints, our method enables training of ultra high resolution images. We demonstrate the effectiveness of our method through superior performance on 7 different benchmarks across classification, object detection, and segmentation. Notably, the proposed method achieves strong performance even on resource-constrained devices like Jetson Nano. Our code is available at https://github.com/Visual-Conception-Group/Localized-Perception-Constrained-Vision-Systems.
\end{abstract}


\section{Introduction}
\label{sec:intro}
The field of image recognition has witnessed tremendous progress in recent years, fueled by groundbreaking architectures like AlexNet \cite{krizhevsky2012imagenet}, ResNet \cite{he2016deep}, ViT \cite{dosovitskiy2020image} and Swin family \cite{liu2022swin}. These architectures excel at processing datasets like ImageNet \cite{deng2009imagenet}, which primarily contain natural images with resolutions well below one megapixel. However, many critical applications rely on the analysis of high-resolution images to extract intricate details about objects of interest. For instance, self-driving cars require the ability to detect distant traffic signs. Similarly, in medical imaging, accurate diagnoses often hinge on examining gigapixel microscope slides that reveal vital information about cells and tissues \cite{huang2022deep}.

While high-resolution images offer undeniable advantages, training deep learning models on such data presents significant challenges. Scaling model architectures to accommodate these massive inputs results in high computational and memory requirements. A seemingly linear increase in the input image size translates to a quadratic rise in computational complexity and memory usage \cite{papadopoulos2021hard}. These requirements can easily create hardware bottlenecks that hinder training.

Existing solutions to tackle this problem such as streaming \cite{pinckaers2020streaming}, gradient checkpointing \cite{marra2020full}, traversal network \cite{papadopoulos2021hard}, iterative patch selection \cite{bergner2022iterative} and Zoom-In \cite{kong2022efficient} primarily address the task of classification. However, the high resolution images pose challenges in various other vision tasks such as object detection and segmentation. And the aforementioned models cannot be extended straightforward for such tasks. Object detection methods such as \cite{ozge2019power}, AdaZoom \cite{xu2021adazoom} and PRDet \cite{leng2022pareto} address the problems of small object detection or training large resolution images. Segmentation methods 
\cite{shen2022high}, \cite{li2024memory} proposed GPU memory-efficient techniques. However, these detection and segmentation methods do not deeply investigate the ability of training very large resolution images under strict GPU memory constraints. It is necessary to study training methods that are capable of training models under such constraints as it helps to train the models even after they are deployed in resource constrained systems. 

\begin{figure}
  \centering
  \includegraphics[width=0.5\textwidth]{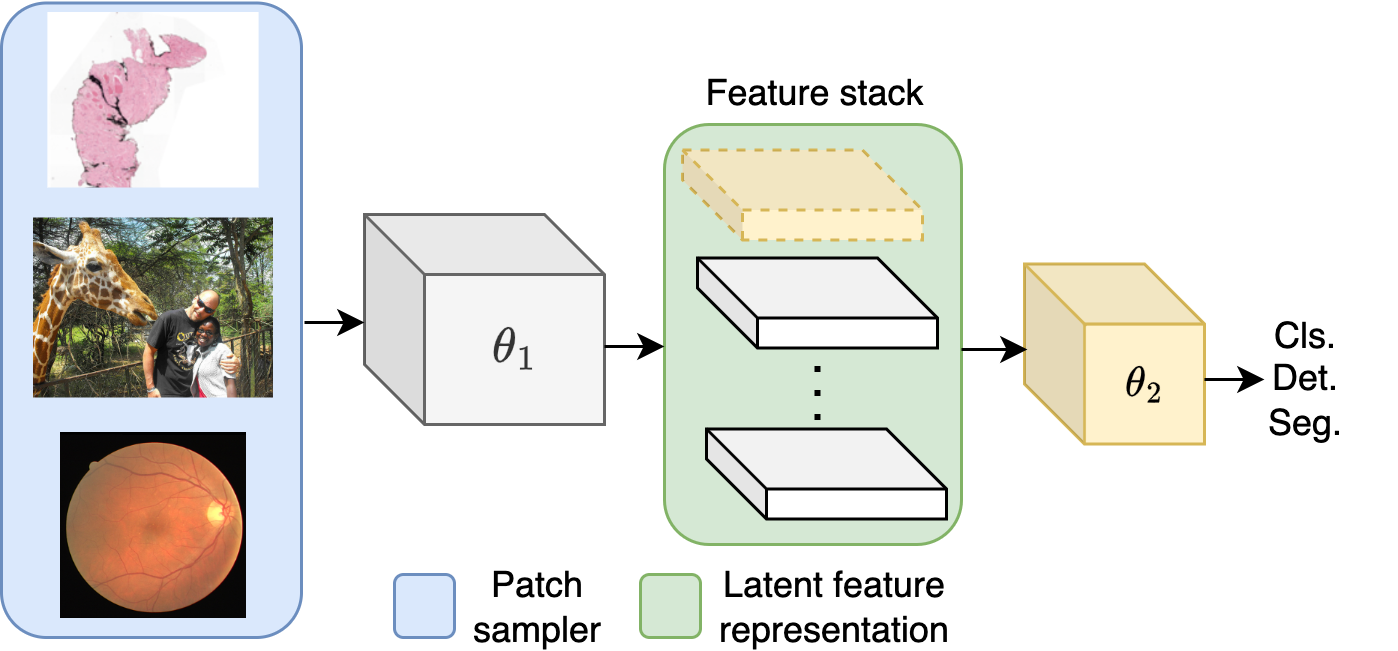} 
  \caption{\small Overview of our framework. It tackles the memory limitations encountered in traditional methods by processing the image in smaller patches. A base model extracts features from each patch and from resized full image. These features are then combined into a latent representation, which essentially captures the essence of the entire image. This latent representation is then processed by an aggregator network to extract both global and fine grained details.}
  \label{fig:illustration}
\end{figure}
In this paper, we propose a framework by leveraging patch based processing \cite{gupta2023patch} alongside global context understanding for training under GPU memory limitations. Our contributions are as follows. \textit{First}, we propose a unified model for image classification, object detection, and segmentation. We use a canonical backbone as feature extractor from local and global patches. An aggregator network then extracts the required fine-grained and global semantics from these patches. \textit{Second}, in our experiments we demonstrate that the patch processing helps in training large images under GPU memory budget. Further, such local and global processing enables capturing fine grained details needed for dense tasks such as object detection and segmentation, and global semantics for classification.
An overview of our model is given in Figure \ref{fig:illustration}. We choose image classification, object detection, and segmentation tasks as they are distinct and the architectures widely differ. Thus a straightforward extension of a method for one task would not easily be applicable to the other task. Further, classification is primarily a global task and needs understanding of the full image. On the other hand, object detection and segmentation are dense tasks, and require both global as well as spatial understanding. 
\section{Related Works}
\textbf{Classification}: CNNs and Transformers have shown tremendous performance for image recognition tasks. While our model can be extended to transformer based models also, we focus on CNNs for brevity. Training these models on high-resolution images remains a challenge due to the substantial memory requirements \cite{bergner2022iterative}. Recent work has focused on addressing this challenge through various approaches.
A common approach to enable high-resolution image processing is to reduce the image resolution before feeding it into the network. This technique is straightforward to implement but can lead to a loss of crucial information, especially when dealing with small objects or fine details. The problem becomes worse when dealing with ultra high resolution images of the dimensions 4096$\times$4096.

Another line of work deals with network architectures for low memory footprint \cite{katharopoulos2019processing, cordonnier2021differentiable}. These architectures often achieve memory efficiency by using techniques like depthwise separable convolutions or by reducing the number of channels in the network. However, these approaches may come at the cost of reduced accuracy compared to standard CNN architectures. 
PatchGD \cite{gupta2023patch} uses a base network to extract features from non-overlapping patches and then pools them using another lightweight network. Attention mechanisms have recently emerged as a powerful tool for improving the efficiency of neural networks \cite{vaswani2017attention}. These mechanisms allow the network to focus on the most relevant parts of the input image. However, attention mechanisms can themselves be computationally expensive, especially for high-resolution images. \\

\noindent{\textbf{Detection}}: Prior literature address small object detection on very large images \cite{liu2023coarse}. 
In a related line of work, in order to fuel the progress of small object detection, datasets such as DOTA \cite{ding2102object}, SODA-A and SODA-D \cite{cheng2023towards} have also been proposed.
\cite{ozge2019power} addresses the issue of small object detection in high resolution images under limited memory constraints. While this is close in spirit to the proposed work, the differences are twofold. First, \cite{ozge2019power} is designed for pedestrian and vehicles only, whereas, our method has no such limitations. Second, \cite{ozge2019power} processes each tile independently and does not get a global picture, whereas, our method extracts global comprehension via learning as well as using a resized version of the full image. AdaZoom \cite{xu2021adazoom} and PRDet \cite{leng2022pareto} networks innovate in object detection within large-scale and drone-view scenarios, focusing on overcoming challenges posed by small objects and scale variation. 
QueryDet \cite{yang2022querydet} and SDPDet \cite{yin2024sdpdet} also address similar challenges.\\ 

\noindent{\textbf{Segmentation}}: Ultra-high resolution image segmentation poses a significant challenge in applications such as Unmanned Aerial Vehicles (UAVs) due to limited computational resources. With various tasks like mapping and decision-making, the need for memory-efficient models becomes crucial. As global segmentation necessitates a lot of memory requirement, some works exclusively address this problem using local patches \cite{chen2019collaborative, wu2020patch, huynh2021progressive, li2021contexts, guo2022isdnet}. Dataset such as URUR with ultra high resolution images (5120$\times$5120) \cite{ji2023ultra} have also been introduced to evaluate the segmentation performance. In case of generic image segmentation, \cite{shen2022high} Shen \textit{et. al.} propose Continuous Refinement Model (CRM) for ultra high-resolution image segmentation refinement. 
\cite{li2024memory} proposes a GPU memory-efficient framework to tackle ultra-high resolution image segmentation. \\

\noindent{\textbf{Preliminaries}}:
As our approach is inspired from patch based processing, we briefly review patch based classification technique proposed in PatchGD \cite{gupta2023patch} algorithm. Instead of updating the entire image at once, PatchGD operates on small image patches, hypothesizing that this localized approach can yield effective model updates while efficiently covering the entire image over iterations. Further, such patch processing can also help training on very large images. Central to PatchGD is the construction of a deep latent representation, denoted as $Z\in \mathbb{R}^{m\times n\times d}$ block, where $mn$ is the total number of patches and $d$ denotes the feature dimensionality.
$Z$ encapsulates information from different parts of the image obtained during previous update steps. $Z$ block serves as an encoding of the full input image and is generated by processing image patches independently using a base model parameterized with weights $\theta_1$. 

To pool the features, a small subnetwork $\theta_2$ is used. Importantly, the computational cost of integrating this subnetwork is minimal. During training, both model components $\theta_1$ and $\theta_2$ are updated. The PatchGD method overcomes limitations of traditional gradient descent methods when handling large images, by avoiding updates on entire image samples and instead computing gradients on localized patches. This approach enhances memory efficiency and computational scalability, making it well-suited for processing high-resolution images in classification.
\begin{figure}
  \centering
  \includegraphics[width=8cm]{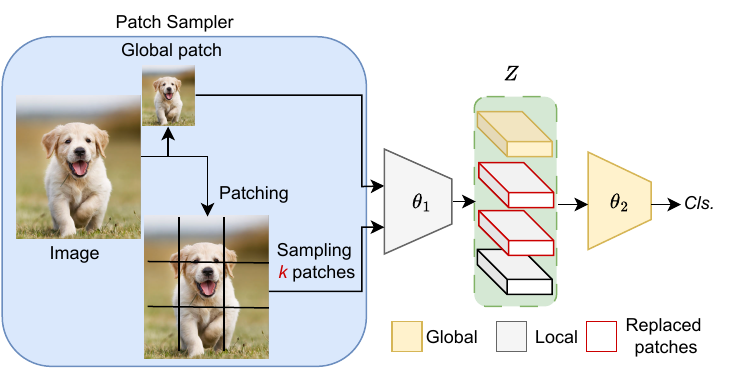}
  \caption*{(a) Classification.}
  \label{fig:Classification}

  \includegraphics[width=8cm]{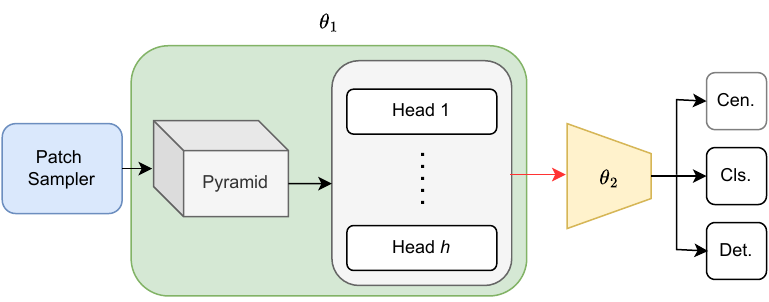}
   \caption*{(b) Detection.}
  \label{fig:Detection}
  
  \includegraphics[width=8cm]{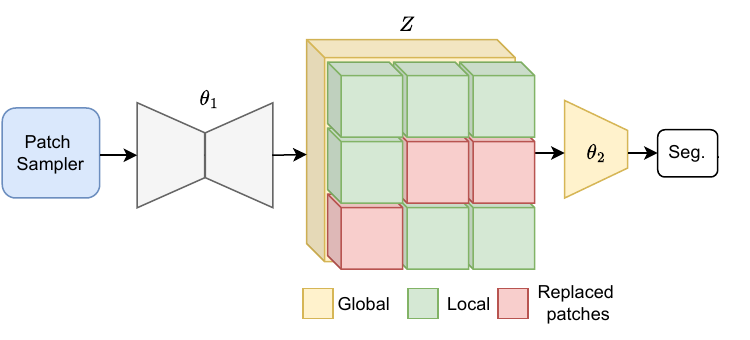}
   \caption*{(c) Segmentation.}
  \label{fig:Segmentation}
  \caption{Block diagram of our method. Patch sampler samples $k$ patches. The features of these patches, obtained from $\theta_1$, are then used to replace the features of latent feature representation $Z$. The feature of the global patch is fused with the patch features. Aggregated features $Z$ are then pooled using model $\theta_2$. The updated $Z$ is used to compute loss and the gradients are computed wrt. the updated features only. The gradients are saved after every inner iteration and gradient accumulation is further used to efficiently consume GPU memory.}
  \label{fig:model}
\end{figure}

\section{Proposed Methodology}
In the following we describe our method for classification, detection and segmentation. Our method is inspired from PatchGD \cite{gupta2023patch}. However, PatchGD only addresses the task of classification. As opposed to classification, object detection and segmentation also need better spatial understanding. Thus, our method addresses this challenge.
\subsection{Image Classification}
\label{sec:class}
Let $X \in \mathbb{R}^{M\times N\times c}$ denote a $c$ channel input image of size $M\times N$. We first divide $X$ into non-overlapping patches of size $M/m\times N/n\times c$ and obtain features using $\theta_1$. In a constrained memory scenario, using all patches is not feasible and will be equivalent to using the full image resolution in terms of compute resources. To efficiently process the image via patches, only a subset of $mn$ patches are sampled at random. This is referred to as sampling in inner iteration. 
These features are then combined to obtain a latent representation $Z \in \mathbb{R}^{m\times n\times d}$, where $d$ is the dimensionality of the feature obtained from each patch. At every inner iteration, only $k$ patches are updated in $Z$. Thus at the end of $J$ inner iterations, the model only sees an equivalent of $kJ$ patches of a given image. 

Even though only subsets of patches are updated in any given iteration, the
$Z$ block accumulates and retains features from all sampled patches. This accumulated representation allows the model to maintain a global perspective and ensure continuity across different parts of the image. It enables effective learning from large-scale data without the need for the entire image to be present in memory at once.
PatchGD aggregates the features obtained from different patches to infer global semantics. However, we find that for very large images such as 4096$\times$4096, it 
does not sufficiently capture the global context. Further, the global semantic inference is also heavily contingent on the sampling rate. If we increase the sampling rate, the memory size increases which makes it difficult to train on the resource constrained devices. 

To address the aforementioned limitation, we use a global context. This is achieved by using a downsampled version of full image as a patch along with other patches. We call this patch as global patch. The global patch is of dimension $M/m\times N/n\times c$. The feature vector $g$ of the global patch, obtained via $\theta_1$, is fused with feature of other patches to obtain a joint representation $Z_g = Z + g$. We then pass $Z_g$ to $\theta_2$ to obtain the classification label. The training objective is given by $\mathcal{L}_{cls} = \mathcal{L}_{CE}(X,p_j, \theta_1, \theta_2)$. $\theta_1$ and $\theta_2$ denotes the model parameters, $p_j$ denotes the patches sampled from image $X$ during $j^{th}$ inner iteration. Since only $k$ patches are updated in $Z$, the gradients are computed with respect to those updated samples only. This helps in efficient usage of GPU memory. 
It is noteworthy that there are other ways of fusing $Z$ and $g$ such as concatenation. We find that addition gives better results for the task of classification. The block diagram of our method is provided in Figure \ref{fig:model}(a). 

\subsection{Object Detection}
 We consider a classical object detector as the base model $\theta_1$. For instance,
FCOS \cite{tian2019fcos}, an anchor-free object detection algorithm can be used as our base model.  
There are $mn$ local patches and a global patch. We feed these patches separately to $h$ heads. With $h$ heads of the detector and $mn +1$ patches, we obtain a feature representation for an individual head as $Z \in \mathbb{R}^{(mn + 1)\times H\times W\times d}$. The updation of $Z$ is similar to the updation described for classification in Section \ref{sec:class}. $\theta_1$ contains the backbone, pyramid and heads. For each of the head output, we reshape the dimensions as $H\times W\times d(mn+1)$. The reshaped feature is then fed to $\theta_2$ which contains a single CNN layer. 
After this step, there are $h$ outputs, of dimensions $H\times W\times d$. It is then passed to the final layer of classification, centerness and regression to get the actual bounding boxes and labels, similar to the method followed in \cite{tian2019fcos}. We train the network using the FCOS \cite{tian2019fcos} objective which includes focal loss \cite{lin2017focal} and IOU loss \cite{yu2016unitbox}.  The block diagram of our method is provided in Figure \ref{fig:model}(b).


\subsection{Segmentation}
Similar to classification, we consider a classical segmentation model as the base model $\theta_1$. For instance, U-net can be used as our base model. 
Each patch is fed as an input to the model $\theta_1$. The output is the form of $M/m\times N/n \times C$. The respective features of all $mn$ patches are then tiled to create a joint representation of the image with dimensions $Z \in \mathbb{R}^{M\times N\times C}$. $Z$ is updated similar to the updation in classification in Section \ref{sec:class}. 
This representation is too heavy to be processed in GPU memory when the resources are constrained. In order to reduce the size of latent representation, we replace the classification layer of $\theta_1$ with a 1$\times$1 convolution. We choose output channels as 8 to be able to process it in limited memory case. The joint representation is then passed through $\theta_2$ to learn the global context. In order to further provide global context, we use the global patch and obtain its representation via $\theta_1$. We further upsample it to $M\times N\times C$ and concatenate with the patch representation such that $Z_g \in \mathbb{R}^{M\times N\times 2C}$. Here, $\theta_2$ comprises of 1$\times$1 convolution to obtain the final mask. Dice loss and BCE (Binary Cross Entropy) is used as training objective. Since the object sizes can be very small as compared to background, Dice loss allows us to handle the class imbalance very well.

\section{Experiments}
The implementation of the algorithms is done in PyTorch. We use AdamW as the optimizer. In case of classification and detection, we use 1e-3 as the initial learning rate. We limit the batch size to 4 for training the models. We also add learning rate warmup and linear decay using methodology proposed in \cite{ma2021adequacy}. In case of segmentation, we use a learning rate of 1e-4. In our experiments, we demonstrate that when the models are trained under the constraint of GPU memory, our method outperforms the baseline by a significant margin. Further, this observation implies that our method can be trained on devices with less memory by using small resolution patches while matching the model' performance on higher resolution images. To further address memory constraints, we use gradient accumulation. At every inner iteration we compute gradients and save it. We set gradient accumulation steps to a maximum of 3 for classification, 10 for detection and 2 for segmentation. 
\subsection{Image Classification}
The experiments for our proposed classification algorithm are conducted on Prostate cANcer graDe Assessment (PANDA) \cite{bulten2022artificial} and UltraMNIST \cite{gupta2022ultramnist}. The number of epochs is 100. 
\begin{table}[ht]
\begin{tabular}{|l|l|l|l|l|l|l|l|}
\hline
{Model} & {Res.} & {Patch} & {S(\%)} & GB  & {Iter.} &  GB(peak) & {Acc} \\
\hline\hline
\cite{gupta2022ultramnist}$^{*}$ & 512 & 128 & 20 & 16 & 4  & 10 & 53.4\\
& 2048 & 128 & 7 & 16 & 11  & 12 & 59.6\\
&4096& 128 & 7 & 24 & 11  & 20 & 65.3\\
&4096 & 256 & 7 & 24 & 11 & 20 & 64.5\\
\hline
Ours & 512 & 128 & 20 & 16 & 4  & 10 & 59.3\\
 & 2048 & 128 & 7 & 16  & 11& 13 & 63.9\\
 & 4096 & 128 & 7 & 24 & 11& 21 & 69.2\\
 & 4096& 256 & 7 & 24 & 11  & 21 & 71.4\\
\hline
\end{tabular}
\caption{\small Classification on PANDA.}\label{table:PANDAResults}
\end{table}%

\begin{table}[ht]
\begin{tabular}{|l|l|l|l|}
\hline
{Method} & {Model} &  {GB(peak)} & {Acc.} \\
\hline\hline
SGD & RN50 & -- & 52.9\\
& MNV2 & -- & 67.1\\ \hline
\cite{gupta2022ultramnist}$^{*}$ & RN50 & 2.5 &62.1\\ 
 & MNV2 & 2.3 & 72.4\\ \hline
Ours &  RN50 & 2.7 & 63.3\\ 
&  MNV2 & 2.6 & 73.5\\
\hline
\end{tabular}
\caption{\small UltraMNIST results.}\label{table:UltraMNISTResults}
\end{table}

\textbf{PANDA}:
 We train our proposed algorithm on PANDA dataset on 24GB and 16GB memory. We use 8,616 samples for training, 1,000 samples for validation, and 1,000 samples for inference, and report the results of inference in Table \ref{table:PANDAResults}. ResNet50 \cite{he2016deep} is used for both 
$\theta_1$ and $\theta_2$. We consider different image sizes of 512$\times$512, 2048$\times$2048 and 4096$\times$4096. The patch size is 128$\times$128 or 256$\times$256.  In case of 512$\times$512, our method scores an accuracy of 59.3\%, whereas, PatchGD achieves 53.4\%. Note that the exact split is not mentioned in \cite{gupta2023patch} and therefore we reproduce the results (indicated by $^*$). In case of very high resolution 4096$\times$4096 with a patch size of 256$\times$256, our method outperforms PatchGD by 6.9\%. This is attributed to the fact that we provide a global context using a global patch. We also note that the peak memory (GB(peak)) consumed by our method for this case is 21GB, whereas, for PatchGD it is 20GB. Thus, there is a slight increase of memory usage in our case but it gives huge boost in terms of accuracy. For a 512$\times$512 image, with a patch size of 128$\times$128, there are 16 patches. With a 20\% sampling rate (S in the Table), 3 patches are randomly selected per inner iteration (indicated by Iter.), totaling 12 patches over 4 iterations. 

\textbf{UltraMNIST}:
Here, we train our algorithm on Jetson Nano (4GB memory). We use 22,400 samples for training, 2,800 samples for validation and 2,801 samples for inference. Image resolution is 512$\times$512, patch size is 256$\times$256, inner iterations is 3 and sampling rate is 25\%.
We use ResNet50 (RN50) as well as MobileNetV2 (MNV2) \cite{sandler2018mobilenetv2} for the two networks and report the results in Table \ref{table:UltraMNISTResults}. We would like to emphasize that Jetson Nano has limited GPU memory and as such training using 512$\times$512 may be unstable. This is seen when SGD is applied.
SGD with ResNet50 gives about 52.9\%, whereas, gives 67.1\% with MobileNetV2. On the other hand PatchGD achieves 62.1\% while using ResNet50. In contrast, we see a boost of 1.2\% in our case. We see a similar observation with MobileNetV2. Further, similar to the observation in PANDA, we see a minor increase in peak memory usage. 

\begin{table*}[ht]
\begin{center}
\begin{tabular}{|l|l|l|l|l|l|l|l|l|}
\hline
{Method} & {Res.}  & {$AP$} & {$AP_{50}$} & {$AP_{75}$} & {$AP_S$} & {$AP_M$} & {$AP_L$} &  GB(peak)\\
\hline\hline
CenterNet$^*$ \cite{duan2019centernet} & 511$\times$511  & 20.2 & 42.6 & 20.4 & 13.1 & 16.3 & 36.8 & 12\\
FCOS$^*$\cite{tian2019fcos} & 800$\times$1024  & 21.5 & 45.3 & 20.5 & 12.3 & 18.4 & 37.2 &  15\\
CornerNet$^*$ \cite{law2018cornernet} & 511$\times$511  & 21.9 & 43.7 & 20.8 & 14.1 & 19.4 & 38.2 &  12\\
YOLOX$^*$ \cite{ge2021yolox} & 640$\times$640  & 23.3 & 46.4 & 21.7 & 15.8 & 28.3 & 40.3 &  17\\
Ours & 4096$\times$4096 & 25.4 & 49.2 & 25.2 & 17.4 & 29.2 & 42.6 & 20\\
\hline
\end{tabular}
\end{center}
\caption{\small Object detection on COCO on 22GB memory. Patch size is 512$\times$512.}\label{table:COCO2017Results}
\end{table*}
\subsection{Object Detection}
Experiments for the proposed detection algorithm are conducted on COCO dataset \cite{lin2014microsoft}, Small Object Detection dAtaset under Driving scenarios (SODA-D) \cite{cheng2023towards} and TJU-Ped-traffic \cite{pang2020tju} in supplementary. We use FCOS \cite{tian2019fcos} as $\theta_1$. We trained the models for 300 epochs. Two Nvidia 2080ti GPUs (11GB memory each) are used for training. \\

{\textbf{COCO}}:
We follow the protocol given in \cite{tian2019fcos}. 
Since image size is 640$\times$480, we create a blank image of size 4096$\times$4096 and add 36 images to it to train and test the model on these large synthesized images. We also resize our new synthesized images to 800$\times$1024 equivalent to the image size in \cite{tian2019fcos} and train FCOS algorithm. We also train CenterNet, CornerNet and YOLOX on the new synthesized images downscaled to the resolution mentioned in Table \ref{table:COCO2017Results}.
We present our results in Table \ref{table:COCO2017Results}. In terms of $AP$ our model obtains 25.4\% compared to 21.5\% of FCOS. Our model adds a slight overhead of about 3.1\% in terms of number of parameters. Due to the patch based processing, we also see that the latency increases. YOLOX obtains a score of 23.3\%. We see a similar trend for other metrics. The results indicate that for given GPU memory, our training method is more efficient as it can capture both fine-grained as well as global details.

 In terms of peak memory usage, our method uses 20GB, YOLOX uses 17GB and FCOS uses 15GB. Though we incur an additional 5GB compared to FCOS, we can see that the proposed method allows training on large images while also improving the performance.  \\

\begin{figure*}
\begin{center}
\begin{tabular}{ccc}
{{{
\includegraphics[width=5.5cm]{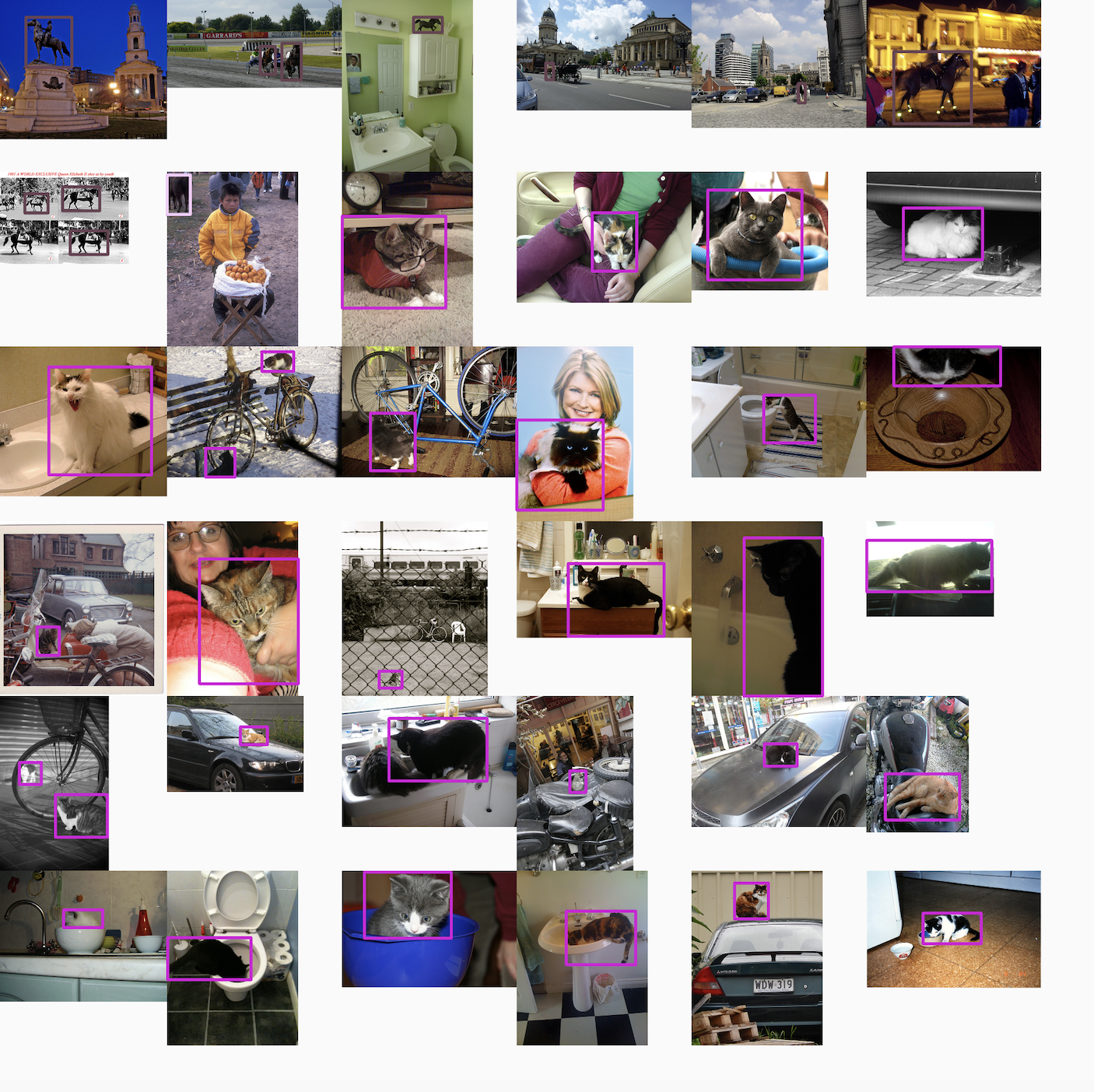}}}}&
{{\includegraphics[width=5.5cm]{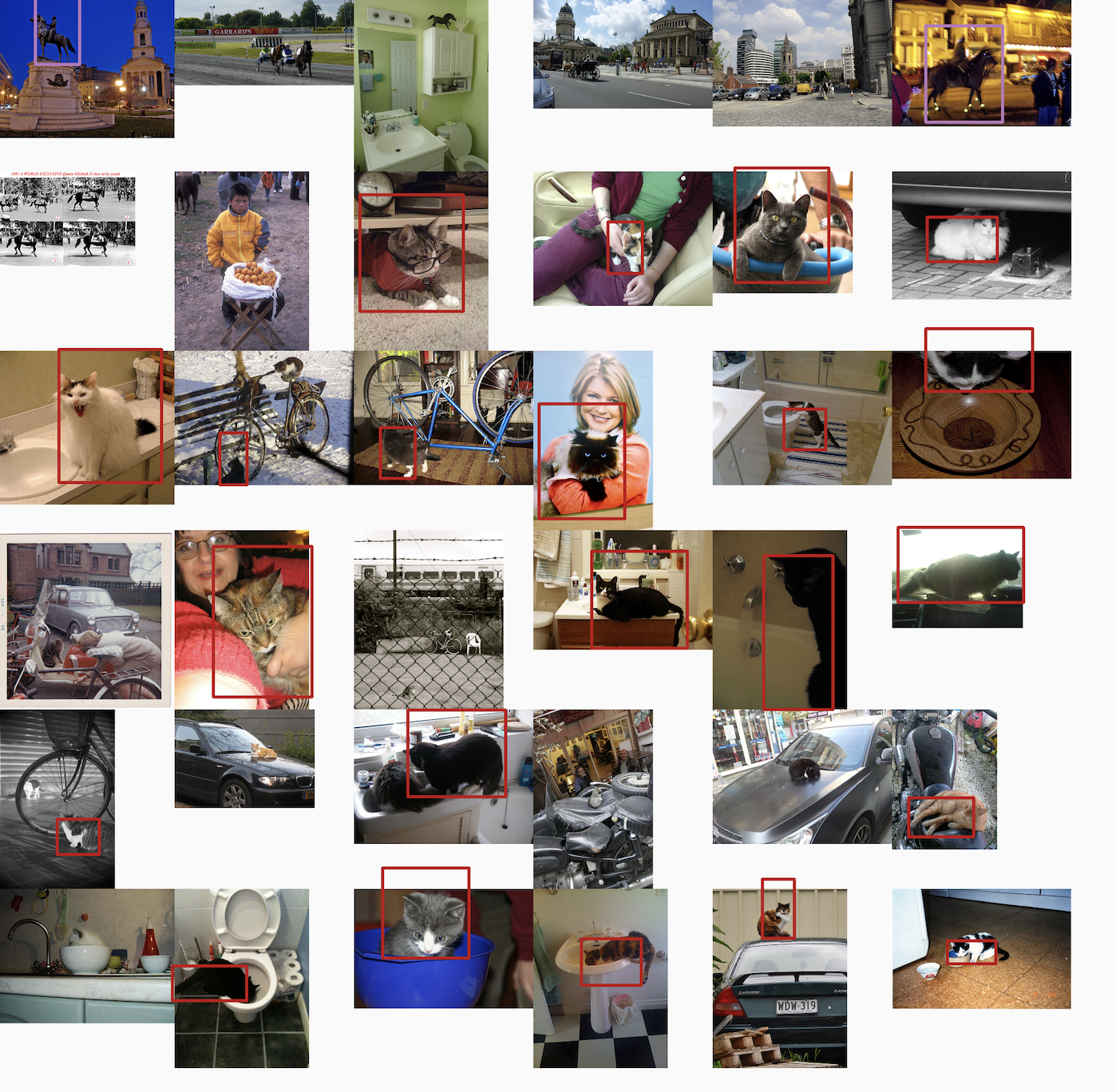}}}&
{{\includegraphics[width=5.5cm]{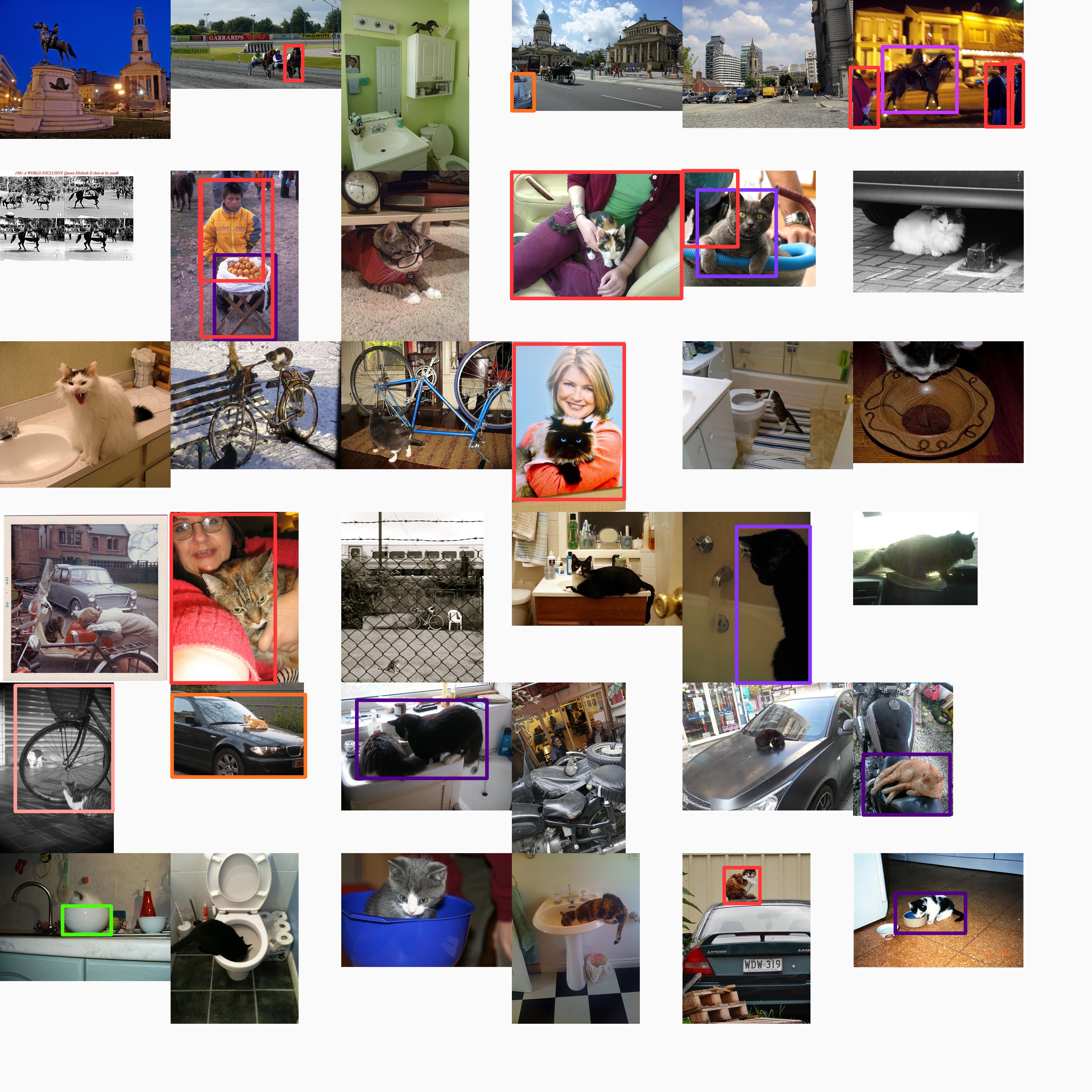}}}\\
(a) Ours &(b) FCOS&(c) YOLO-X
\end{tabular}
\end{center}
\caption{Predicted bounding boxes for COCO synthesized images. Resolution is 4096 $\times$ 4096. }
\label{fig:predicted-box}
\end{figure*}

 \textbf{Predicted Bounding Boxes}
The predicted bounding boxes using our proposed object detection algorithm for new synthesized images are given in Figure \ref{fig:predicted-box}(a). We also give the predicted bounding boxes using FCOS and YOLOX in Figures \ref{fig:predicted-box}(b) and \ref{fig:predicted-box}(c) respectively. It can be inferred that FCOS fails to detect a lot of small and medium sized objects when the image size is large and YOLOX performs well but fails to outperform the proposed model.

\begin{figure*}
\begin{center}
\begin{tabular}{ccc}
{{{
\includegraphics[width=5.5cm]{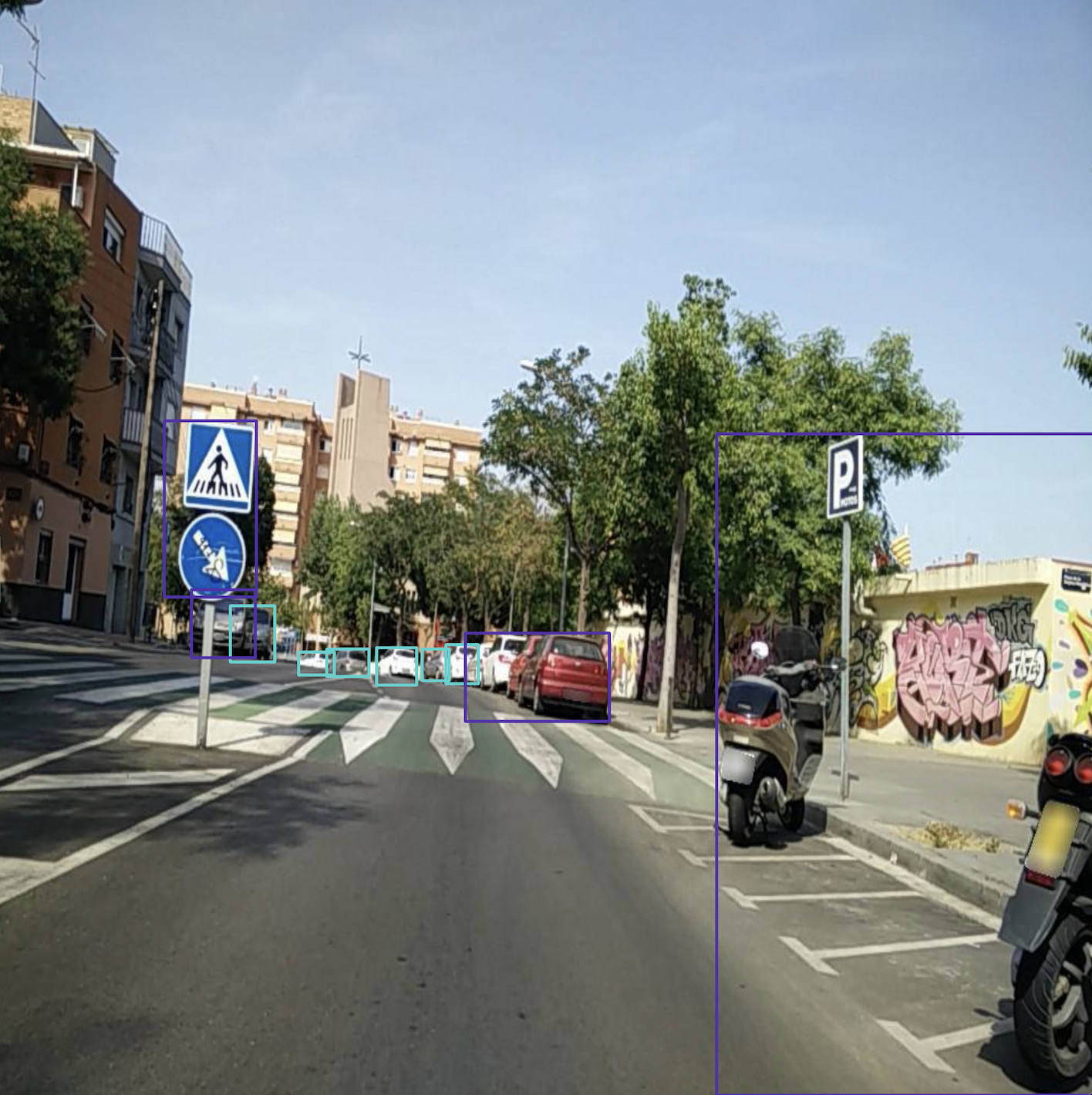}}}}&
{{\includegraphics[width=5.5cm]{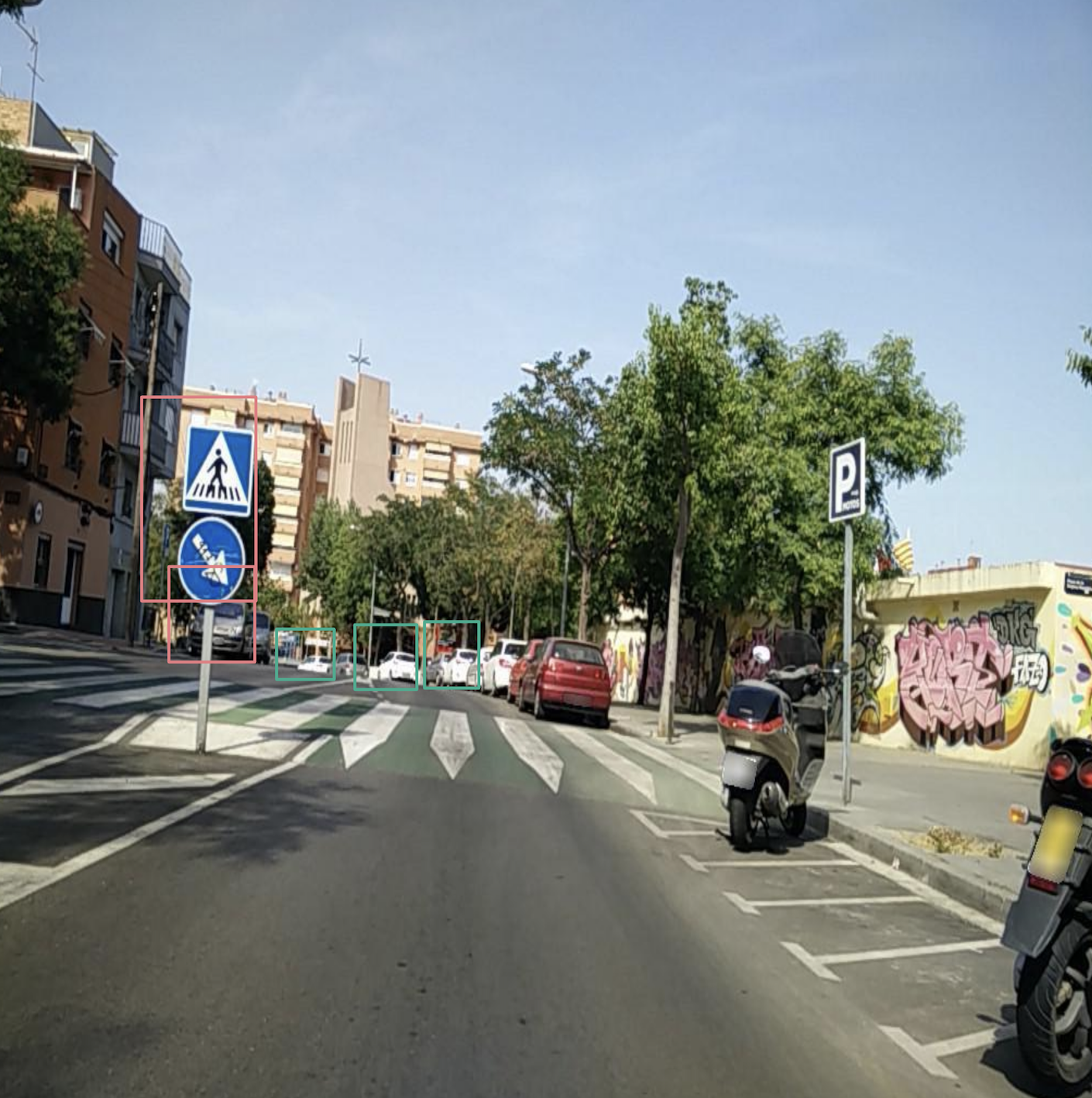}}}&
{{\includegraphics[width=5.5cm]{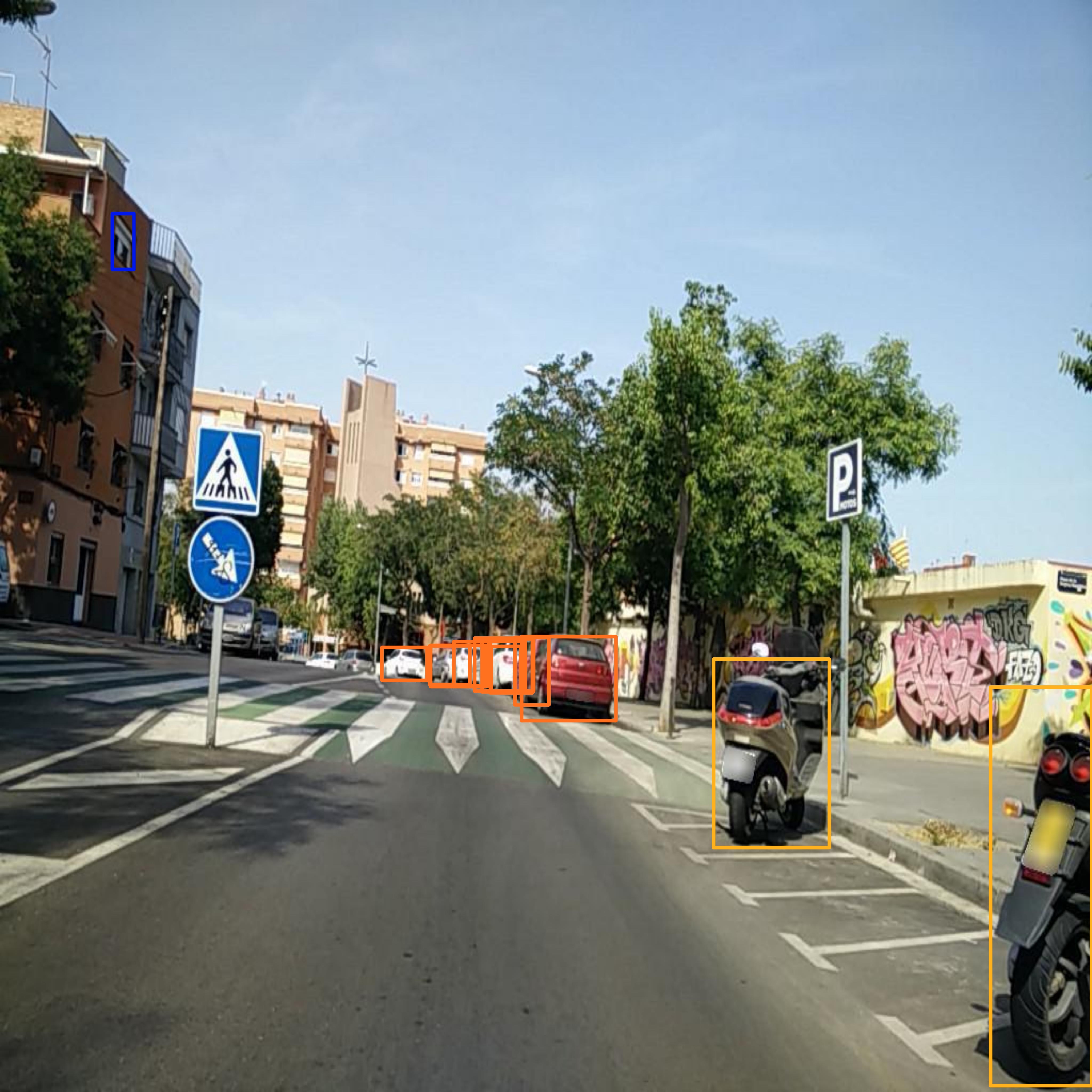}}}\\
(a) Ours &(b) FCOS&(c) YOLO-X
\end{tabular}
\end{center}
\caption{Predicted bounding boxes on SODA-D. Resolution is 4096 $\times$ 4096}
\label{fig:predicted-box-soda}
\end{figure*}

{\textbf{SODA-D}}: We follow the protocol in \cite{cheng2023towards}. The average image size is 3407$\times$2470. We resize the images to 4096$\times$4096. We present our results in Table \ref{table:SODADResults}. $^\dagger$ indicates that the results 
\begin{table}
\begin{center}
\begin{tabular}{|l|l|l|l|l|}
\hline
{Method}  & {$AP$} & {$AP_{50}$} & {$AP_{75}$} & {Par.} \\
\hline\hline
FCOS\textsuperscript{\textdagger} & 23.9 & 49.5 & 19.9 & 32 \\
CornerNet\textsuperscript{\textdagger}  & 24.6 & 49.5 & 21.7 & 201  \\
YOLOX\textsuperscript{\textdagger}& 26.7 & 53.4 & 23.0 & 9  \\
Ours  & 28.6 & 54.2 & 23.3 & 33 \\
\hline
\end{tabular}
\end{center}
\caption{\small SODA-D detection results. } \label{table:SODADResults}
\end{table}
  are reported from \cite{cheng2023towards}.
Our model achieves an $AP$ of 28.6\% compared to 23.9\% of FCOS and 26.7\% of YOLOX. Our method outperforms other methods in terms of $AP$, $AP_{50}$ and $AP_{75}$. It is noteworthy that \cite{cheng2023towards} runs the detector on patches of resolution 1200$\times$1200 and the patch-wise detection outcomes are mapped back to the original images, followed by applying Non Maximum Suppression to eliminate overlapping and redundant predictions. This is an efficient trade-off between using GPU memory and processing very high resolution images. Our method uses a full image resolution of 4096$\times$4096 with a patch size of 512$\times$512. Peak memory usage is 18GB for our method. \cite{cheng2023towards} uses 4 NVIDIA GeForce RTX 3090 GPUs to train the models. From Table \ref{table:SODADResults} it can be inferred that the proposed method processes the high resolution images under memory constraints with a better trade-off.

The results for predicted bounding boxes are given in Figure \ref{fig:predicted-box-soda}.

\textbf{TJU-Ped-traffic}
We follow the protocol given in \cite{pang2020tju}. We use 13,858 samples for training, 2,136 samples for validation, and 4,344 samples for inference. It has 52\% images with a resolution of 1624$\times$1200 and 48\% images have a resolution of at least 2560$\times$1440.
We present our results in Table \ref{table:TJUDHDResults}. Our method obtains an $AP$ of 29.45, YOLOX achieves 27.2\%, and FCOS obtains 23.3\%. We see a similar trend for other metrics also.
We also note that our method' peak memory is 20GB compared to 18GB of YOLOX and 14GB of FCOS. This indicates that the proposed method can process full resolution images under memory constraints and also give better performance.

\begin{table*}[ht]
\begin{center}
\begin{tabular}{l|l|c|c|c|c|c|c|c|c}
\hline
{Method} & {Res.}  & {$AP$} & {$AP_{50}$} & {$AP_{75}$} & {$AP_S$} & {$AP_M$} & {$AP_L$} & {M} &  {GB}\\
\hline
CenterNet$^*$ \cite{duan2019centernet} & $511 \times 511$ & 24.5 & 44.7 & 21.4 & 15.6 & 18.3 & 39.1 & 71 &  12\\
FCOS$^*$\cite{tian2019fcos} & $800 \times 1024$ & 23.3 & 44.9 & 22.5 & 15.3 & 16.4 & 38.3 & 32 &  14\\
CornerNet$^*$ \cite{law2018cornernet} & $511 \times 511$  & 24.9 & 45.9 & 23.3 & 15.4 & 18.9 & 39.4 & 201 &  13\\
YOLOX$^*$ \cite{ge2021yolox} & $640 \times 640$ & 27.2 & 47.7 & 25.1 & 17.6 & 20.3 & 41.8 & 9 &  18\\
Ours & $4096 \times 4096$  & 29.4 & 49.8 & 26.6 & 19.1 & 22.7 & 43.2 & 33 &  20\\
\hline
\end{tabular}
\end{center}
\caption{Results for object detection on TJU-Ped-traffic dataset on 22GB memory. Patch size is 512 $\times$ 512. M denotes the parameters in millions. GB is the peak memory used.}
\label{table:TJUDHDResults}
\end{table*}

\subsection{Segmentation} We use U-net \cite{ronneberger2015u} and DeepLab v3 \cite{chen2017rethinking} as $\theta_1$. The models are trained for 100 epochs on 3090Ti GPU (24GB memory) as well as on Jetson Nano. 

\textbf{DRIVE - Retina Blood Vessel}:
DRIVE \cite{staal2004ridge} dataset has a total of 100 images. Each image is of resolution 512$\times$512. As per the protocol in \cite{abdallahvesselseg}, we split the dataset into 80 for training and 20 for test. 
In Table \ref{table:DRIVEresults}, U-net and DeepLab v3 with Type Full are trained on full image of resolution 512$\times$512. At the time of inference also, full resolution is used.
U-net and DeepLab v3 with Type Downsampled (Down.) are trained with the image downsampled to 128$\times$128. At the time of inference, the input is resized to 128$\times$128 and the output mask is upsampled to 512$\times$512. Ours$^\ddagger$ indicates that the model is trained with crops of size 128$\times$128 under the proposed framework without including the global patch. Finally, Ours indicates training with crops of size 128$\times$128 along with the global patch.

\begin{figure*}[ht]
\begin{center}
\begin{tabular}{cccccc}
{{{\includegraphics[width=2.5cm, trim={5cm 5cm 5cm 5cm}, clip]{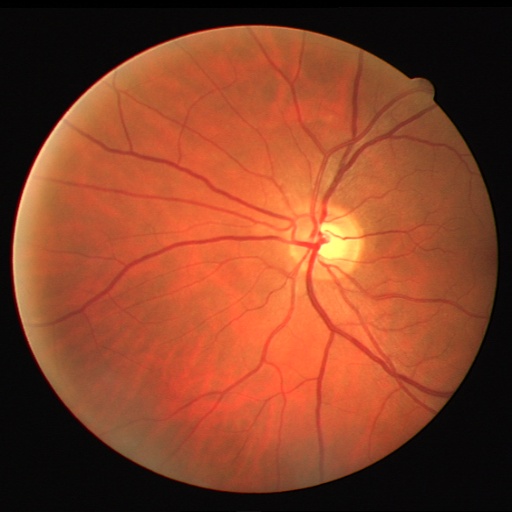}}}}&
{{\includegraphics[width=2.5cm, trim={5cm 5cm 5cm 5cm},clip]{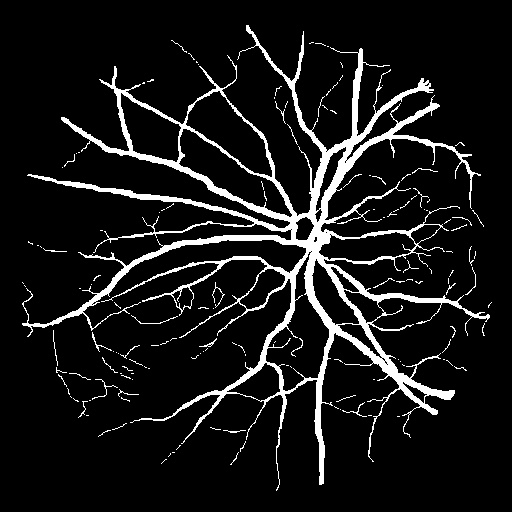}}}&
{{\includegraphics[width=2.5cm, trim={5cm 5cm 5cm 5cm},clip]{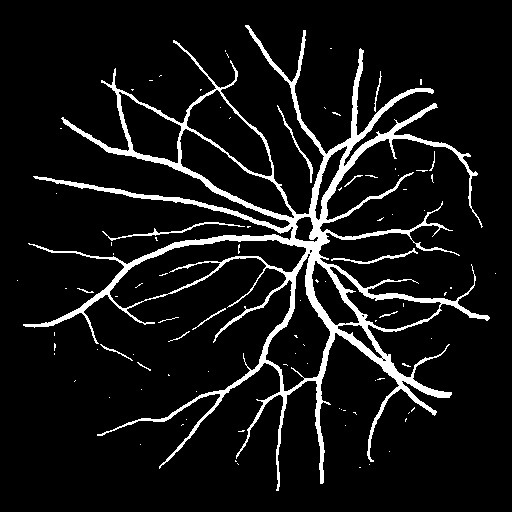}}}&
{{\includegraphics[width=2.5cm, trim={5cm 5cm 5cm 5cm},clip]{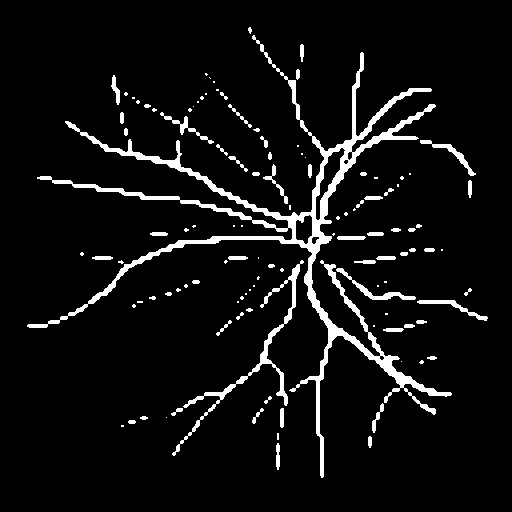}}}&
{{\includegraphics[width=2.5cm, trim={5cm 5cm 5cm 5cm},clip]{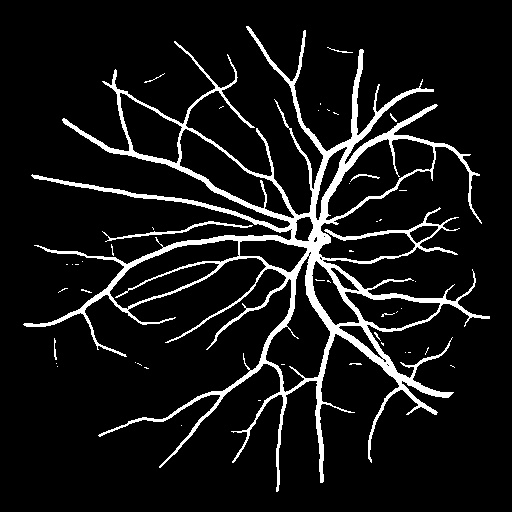}}}&
{{\includegraphics[width=2.5cm, trim={5cm 5cm 5cm 5cm},clip]{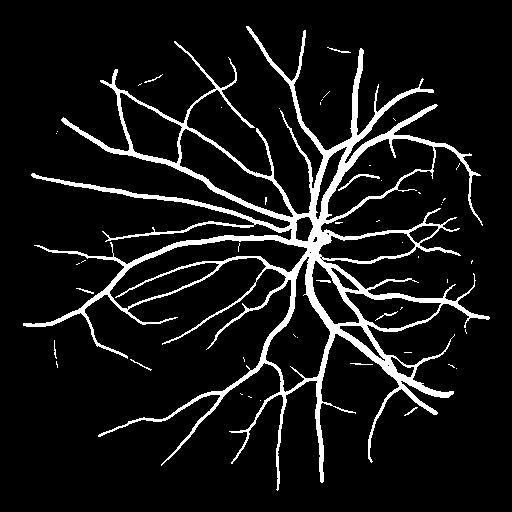}}}\\
(a) IP &(b) GT &(c) Full  &(d) Down.& (e) Ours$^\ddagger$ & (f) Ours
\end{tabular}
\end{center}
\caption{\small Segmented masks for DRIVE image using different Types of Unet. For better clarity, we show only a crop of image.}
\label{fig:seg_image}
\end{figure*}

The Type Full version uses full image and hence occupies more GPU memory. We give it here as a reference. We are particularly interested in comparison with Type Downsampled as it uses only 1.39 GB of GPU memory and hence can be trained within 4GB limit. In terms of F1, IoU, balanced accuracy, PLR and NLR, we see a substantial gain for our method compared to U-net Down. F1 increases from 61.0\% for Type Downsampled to 79.8\% for Ours. IoU increases from 43.9\% to 66.5\%. We can also see that the results for Ours is similar or better than Full.
We also see that the global patch improves the overall performance with a very minor increase in peak GPU usage. We arrive at a similar conclusion using DeepLab v3 as $\theta_1$. We give the qualitative results in Figure \ref{fig:seg_image}. We can see that the continuity and fine details are well preserved in Figure \ref{fig:seg_image}(f).

\begin{table*}[ht]
    \begin{center}
    \begin{tabular}{|c|c|l|c|c|c|c|c|c|c|}
        \hline
        Model & Size & Type & Acc & F1 & IoU & B. Acc & PLR & NLR$\downarrow$ & GB(peak) \\
        \hline\hline
        \rowcolor{gray!30} 
        Unet &    512 &   Full             &  96.3 &  78.2 &   64.2  &   86.7 & 37.7 & 19.2 & 7.0 \\
             &    128 &   Down.            &  93.9 &  61.0 &   43.9  &   75.3 & 16.5 & 31.9 & 1.3 \\
             &    128 &   Ours$^\ddagger$  &  96.4 &  79.4 &   66.0  &   88.5 & 53.3 & 21.3 & 3.4 \\
             &    128 &   Ours             &  96.5 &  79.8 &   66.5  &   89.0 & 50.7 & 20.5 & 3.6 \\
        \hline
        \rowcolor{gray!30}
        DLv3 &   512 &   Full             &  94.0 &  66.3 &    49.6 &   81.9 & 20.3 & 33.8  &  4.5 \\
              &   128 &   Down.           &  89.4 &  33.9 &    20.4 &   63.1 & 6.4  & 72.3 &  1.4 \\
              &   128 &   Ours$^\ddagger$ &  93.9 &  65.0 &    48.2 &   80.8 & 65.9 & 36.2  &  2.8 \\
              &   128 &   Ours            &  93.8 &  65.6 &    48.9 &   81.8 & 64.7 & 33.8  &  2.9 \\
        \hline
    \end{tabular}
    \end{center}
    \caption{\small DRIVE results. B. Acc denotes balanced accuracy and is obtained as arithmetic mean of sensitivity and specificity. $^\ddagger$ denotes without global version.}
    \label{table:DRIVEresults}
\end{table*}

\textbf{Aerial Imagery}:
This dataset consists of 72 images. Each image is of resolution 1024$\times$ 1024. 
We use 52 samples for training and 20 samples for test. We observe that both Ours$^\ddagger$ and Ours perform better than Type Downsampled in terms of accuracy, though lags mildly in other metrics. However, in DeepLab v3 we see that the methods Ours$^\ddagger$ and Ours perform better than Downsampled in terms of accuracy, IoU, balanced accuracy, PLR and NLR. We also perform training on Jetson Nano. The results are reported in Table \ref{tab:segmentation_jetson}. We see that for DeepLab v3 and Unet, training is possible only for downsampled image of size 128$\times$128, whereas, Ours$^\ddagger$ can be trained on 1024$\times$1024. We see a huge increase in all the scores compared to downsampled versions. 
Unlike classification, segmentation is a resource intensive task and we had to make a lot of optimization steps to be able to train on Jetson Nano. We performed additional optimizations like increasing swap memory, switching from OpenCV to PIL and disabling augmentation which allowed successful training. Then leveraging our method, we extended training to 512 and 1024 resolution images. 
\begin{table*}[h]
    \begin{center}
    \begin{tabular}{|c|c|l|c|c|c|c|c|c|c|}
        \hline
        Model & Size & Type & Acc & F1 & IoU & B. Acc & PLR & NLR$\downarrow$ & GB(peak) \\
        \hline\hline
        \rowcolor{gray!30}
        Unet & 1024 & Full           & 95.3 & 38.9 & 29.8 & 67.9 & 22.6 & 37.8 & 21.0 \\
             & 256 & Down.           & 94.9 & 33.8 & 24.6 & 64.8 & 14.9 & 44.0 & 2.2 \\
             & 256 & Ours$^\ddagger$ & 95.1 & 33.7 & 25.4 & 64.9 & 20.7 & 43.9 & 11.2 \\
             & 256 & Ours            & 95.3 & 32.1 & 24.1 & 63.3 & 27.0 & 47.6 & 11.7 \\
             & 128 & Ours            & 95.5 & 32.7 & 25.5 & 64.2 & 23.1 & 45.8 & 3.3 \\
        \hline
        \rowcolor{gray!30}
        DLv3 & 1024 & Full           & 95.9 & 42.4 & 32.8 & 70.0 & 25.1 & 33.4 & 14.6 \\
              & 256 & Down.           & 93.8 & 32.5 & 22.2 & 63.9 & 19.7 & 45.7 & 2.0 \\
              & 256 & Ours$^\ddagger$ & 95.1 & 34.0 & 25.9 & 66.2 & 13.2 & 40.9 & 7.9 \\
              & 256 & Ours            & 95.2 & 31.9 & 24.1 & 64.5 & 21.0 & 44.5 & 8.6 \\
              & 128 & Ours            & 95.2 & 31.3 & 23.2 & 64.1 & 23.2 & 45.4 & 3.7 \\
        \hline
    \end{tabular}
    \end{center}
    \caption{\small Segmentation on aerial dataset.}
    \label{tab:segmentation_metrics}
\end{table*}

\begin{table*}[h]
    \begin{center}
    \begin{tabular}{|c|c|c|c|c|c|c|c|c|c|c|}
        \hline
        Model & Size & BS & Acc & F1 & IoU & B. Acc & PLR & NLR$\downarrow$ & RAM & GB(peak) \\ 
        \hline\hline
        DLv3            & 128  & 5 & 92.2 & 20.4 & 12.7 & 45.4 & 5.8 & 58.2 & 3.6 & 2.1 \\ 
        Unet            & 128  & 5 & 93.3 & 20.3 & 12.7 & 47.3 & 7.8 & 54.1 & 3.8 & 1.7 \\ 
        Ours$^\ddagger$ & 512  & 1 & 94.8 & 26.9 & 19.1 & 48.1 & 19.5 & 53.1 & 3.5 & 1.5 \\ 
        Ours$^\ddagger$ & 1024 & 1 & 95.0 & 33.5 & 25.1 & 52.3 & 20.5 & 44.2 & 3.7 & 2.6 \\ 
        \hline
    \end{tabular}
    \end{center}
    \caption{\small Segmentation of aerial dataset on Jetson Nano. BS is batch size}
    \label{tab:segmentation_jetson}
\end{table*}

\section{Conclusion}
In this work, we address the significant challenge of efficiently processing high-resolution images for tasks like classification, object detection, and segmentation under stringent memory constraints. Our innovative framework integrates localized patch-based processing with a global contextual understanding, enabling comprehensive image analysis under memory constraints. This approach not only preserves the fine-grained details necessary for accurate object detection and segmentation but also incorporates global semantics essential for robust classification performance.
Experiments across seven distinct benchmarks, demonstrates that our method achieves competitive performance. We also demonstrate training on resource-constrained devices, such as the Jetson Nano.


\bibliographystyle{IEEEtran}
\bibliography{energy_efficient}

\end{document}